\documentclass[cameraready]{Interspeech}

\title{I Am No One: Style-Aware Paraphrasing for Text Anonymization}

\author[affiliation={}, orcid=0009-0002-4836-7314, correspondingauthor]{Ahmed Sohair}{Khan}
\author[affiliation={}, orcid=0000-0002-8994-9532]{Estrid}{He}
\author[affiliation={}, orcid=0000-0002-4659-0101]{Monica}{Wachowicz}
\author[affiliation={}, orcid=0000-0001-7640-4624]{Elham}{Naghizade}

\address{
    RMIT University, Melbourne, Australia
}

\email{
    ahmed.sohair.khan@rmit.edu.au,
    estrid.he@rmit.edu.au,
    monica.wachowicz@rmit.edu.au,
    e.naghizade@rmit.edu.au
}

\keywords{ Text anonymization, Authorship attribution, Stylometry, Language processing, Privacy-preserving NLP }

\usepackage{comment}
\usepackage{threeparttable,tabularx,multirow,placeins}

\begin{document}

\maketitle
\begingroup
\renewcommand{\thefootnote}{}
\footnotetext{Code and Supplementary material: \url{https://github.com/ahmedsohair/SAPTA26}}
\endgroup
\begin{abstract}
Authorship attribution models can re-identify users from seemingly anonymized text by exploiting stable stylistic fingerprints, even after explicit identifiers are removed, posing a growing privacy risk for text publishing and analytics. This risk extends to speech-derived text such as ASR transcripts of meetings and call-center conversations, where stylometric leakage can persist even after acoustic anonymization. Differential privacy-based anonymization often severely degrades text quality and utility. We propose a style-aware, prompt-driven anonymization approach that uses pretrained large language models to construct compact stylistic profiles from minimal samples and rewrite text to suppress identifiable style markers while preserving meaning. Across blog and review datasets, our approach reduces authorship attribution F1 by 60--70\% while maintaining content quality and readability, substantially outperforming DP-based and non-DP baselines. 
\end{abstract}

\section{Introduction and Related Work}
The widespread sharing of online content has raised significant concerns about user privacy. Subtle stylistic patterns embedded within user-generated text or ASR transcripts can be leveraged to identify and trace the original author, even when users attempt to remain anonymous. This practice, known as ``authorship attribution,'' poses a threat to individuals' privacy, as it can expose users' identities.

Traditional anonymization techniques, such as removing explicit personal identifiers, are inadequate against modern machine learning models capable of extracting nuanced stylistic cues \cite{lison-etal-2021-anonymisation}. Therefore, text anonymization must also obscure subtle stylistic fingerprints such as syntax, vocabulary, and discourse patterns that enable authorship attribution \cite{sundararajan-woodard-2018-represents}.

Related speech-privacy work has mostly targeted acoustic speaker traits, yet stylometric leakage can persist in speech derived text. Prior studies show de-identification can impact transcription-based depression detection \cite{LopezOtero2017DepressionDU}, and recent work anonymizes \emph{speech content style} via transcript paraphrasing to reduce identification while preserving meaning \cite{Sinha2024SafeguardingSC}.

\begin{figure*}[t]
    \centering
  \includegraphics[width=0.7\linewidth]{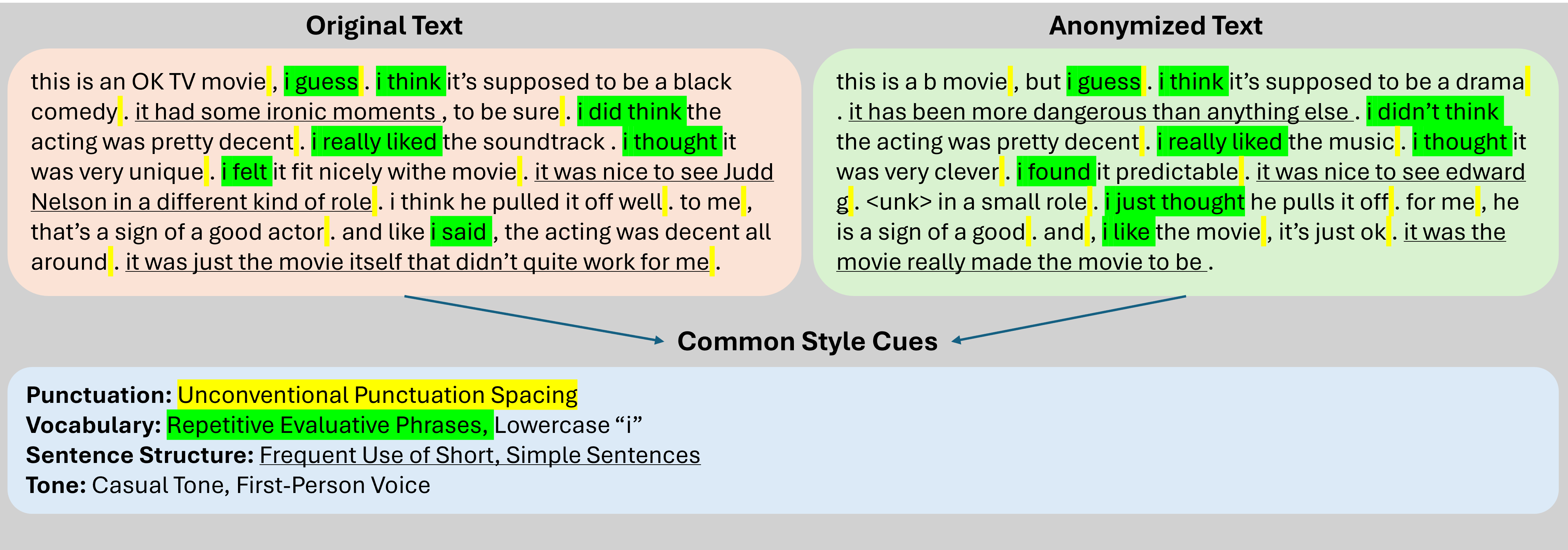}
    \caption{Original text (\emph{left}) and DP anonymized version (\emph{right}). Stylistic cues remain in DP version that may support re-identification.}
    \label{fig:DPexample}
\end{figure*}

Recently, differential privacy (DP)-based methods such as DP-VAE \cite{10.1145/3485447.3512232}, DP-Prompt \cite{utpala-etal-2023-locally}, and DP-MLM \cite{meisenbacher2024dpmlmdifferentiallyprivatetext} have attracted attention for textual anonymization, introducing calibrated noise under privacy budgets. However, DP's mathematical guarantees come with a critical trade-off: the choice of privacy budget ($\varepsilon$) drastically affects utility. High-$\varepsilon$ settings yield poor privacy, while low-$\varepsilon$ settings degrade text to unreadability. Figure~\ref{fig:DPexample} demonstrates that DP-based methods may retain identifiable stylistic patterns despite formal privacy guarantees. This is because latent-level DP mechanisms fail to disrupt higher-order syntactic structures, while token-level noise injection degrades utility without fully erasing authorship cues.

Recent work questions whether rigorous DP is essential for defending against authorship attribution~\cite{meisenbacher2024thinkingoutsidedifferentialprivacy}. Non-DP paraphrasing approaches have emerged, including JAMDEC~\cite{fisher2024jamdecunsupervisedauthorshipobfuscation}, which uses constrained beam decoding, and STYLEREMIX~\cite{fisher-etal-2024-styleremix}, which applies fine-grained style transformations. Other systems, such as ALISON~\cite{xing2024alisonfasteffectivestylometric}, explicitly target stylistic features by ranking style-salient phrases and replacing them, showing that direct stylometric targeting can yield strong privacy gains. Earlier methods such as ER-AE~\cite{bo-etal-2021-er} and DP-Paraphrase~\cite{mattern2022limits} also aimed to balance privacy and utility. Yet most methods still rely on either random perturbations or general paraphrasing without explicit control over stylistic markers.

We posit that effective anonymization should focus on neutralizing author-specific style attributes rather than indiscriminate noise injection. This leads to our central research question: \textit{How can we leverage stylistic transformations to anonymize user-generated text while preserving content utility, without resorting to noise injection?}

We propose a style-aware, prompt-driven anonymization approach that uses large language models to construct explicit stylistic profiles by capturing text length, punctuation patterns, vocabulary choices, and tone. It then rewrites text to obscure these markers while preserving content. Unlike methods relying on noise injection or general paraphrasing, our approach provides fine-grained control to balance anonymization and semantic content preservation. Experiments on two real-world datasets spanning short-form and long-form text show that, by avoiding explicit noise injection, style-focused rewriting can effectively reduce authorship attribution accuracy while maintaining near-original meaning and improving readability.

\section{Methodology}
\label{sec:method}

\begin{figure}[t]
\centering
\includegraphics[width=\linewidth]{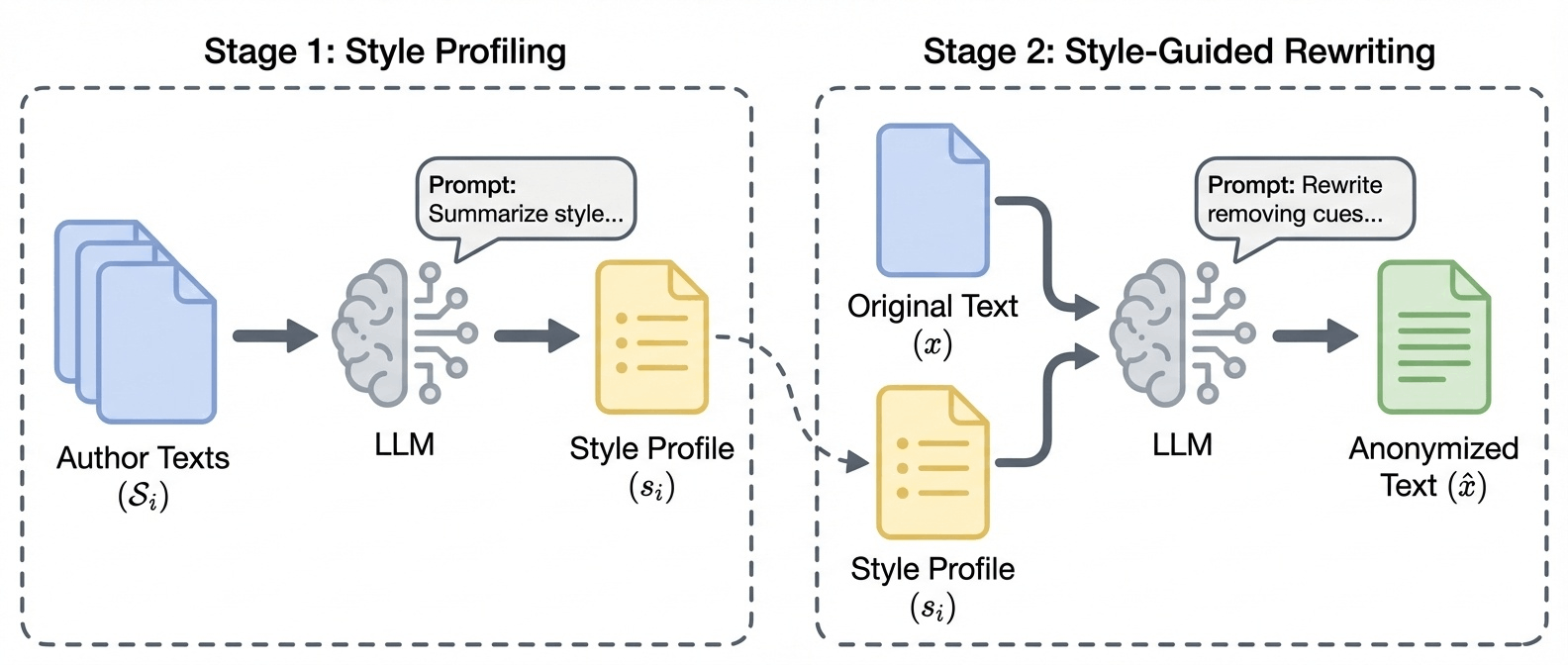}
\caption{(1) Style-Profiling Module generates per-author profiles $s_i$ from $K$ samples; (2) Style-Guided Rewriting Module uses profiles to guide rewriting of text $x$ into anonymized $\hat{x}$.}
\label{fig:methodology}
\end{figure}

Given authors $\mathcal{A}=\{A_1,\dots,A_N\}$ with corpora $\mathcal{D}_i$, we produce anonymized versions $\hat{\mathcal{D}}_i$ by rewriting each input text $x \in \mathcal{D}_i$ into an anonymized output $\hat{x}$. Our goal is that an authorship attribution classifier $\mathcal{F}(\cdot)$ cannot reliably recover the true author. Formally, we aim to ensure $\mathcal{F}(\hat{x}) \neq i$ (or assigns a low probability to author $i$), while preserving semantic content and fluency. Rather than injecting random noise, we reframe anonymization as a \textit{controlled stylistic transformation} problem. Our framework (Figure~\ref{fig:methodology}) consists of two tightly coupled modules: a \textit{Style-Profiling Module} that extracts each author's stylistic fingerprint from minimal samples, and a \textit{Style-Guided Rewriting Module} that uses this profile as a control signal to neutralize those cues.

\subsection{Style-Profiling Module}
\label{sec:style-profiling}

The success of authorship attribution lies in the extraction of distinct stylistic patterns. Informed by classical stylometry, we characterize these fingerprints along four key dimensions: sentence length, vocabulary choice, tone, and punctuation patterns \cite{holmes1994authorship,stamatatos2009survey,usha2017authorship}. Rather than learning implicit style representations, we aim to extract human-interpretable style descriptions for these dimensions.

Given training texts $\mathcal{D}_i^{\text{train}}$ for author $A_i$, we sample $K$ representative texts forming a style evidence set $\mathcal{S}_i = \{x_1, \dots, x_K\}$. We prompt an LLM to summarize the author's writing style across the four key stylometric dimensions, extracting these into a human-readable profile $s_i$. This module has two key controls: (i) \textit{Profile size} $K$ specifies the number of author samples used for style profiling. We find empirically that $K=5$ provides stable, author-distinctive profiles across both datasets, and (ii) \textit{Style channels} controls which stylistic dimensions to include in the profile, i.e., all (full profile capturing all dimensions) or individual dimensions (e.g., Length-only, Tone-only). This enables us to identify the most critical stylistic markers for anonymization.


\subsection{Style-Guided Rewriting Module}
\label{sec:rewriting}

Given source text $x$ and its profile $s_i$, the anonymized output $\hat{x}$ is generated by prompting an LLM to rewrite while suppressing the identified stylistic cues. The prompt instructs preservation of semantic content. Unlike generic paraphrasing baselines that rewrite uniformly, our approach provides fine-grained control: the style profile specifies which dimensions to neutralize, enabling targeted anonymization. This module is \textit{attacker-agnostic}—we assume no access to the adversary's classifier and do not optimize against any specific attacker.

This module has one key control: \textit{Rewrite mode} determines how the style profile guides rewriting. \textit{Style-guided} provides the explicit profile to the LLM (our main approach), enabling precise control over which stylistic markers are suppressed. \textit{Semi-guided} instructs the LLM to rewrite in a neutral style without providing the explicit profile. This aims to test whether explicit profiles improve both privacy and utility. \textit{Unguided} performs generic paraphrasing with no mention of style, serving as a lower-bound baseline.

\section{Experimental Setup}
\label{sec:experimental_setup}

\subsection{Datasets and Baselines}

Following \cite{meisenbacher2024thinkingoutsidedifferentialprivacy}, we use \textsc{Author10}, a subset of the Blog Authorship Corpus \cite{schler2006effects} containing 15,070 long-form blogs from 10 authors. For short-form evaluation, we construct \textsc{Illinois9}, a subset of Google Reviews \cite{li-etal-2022-uctopic} with 3,959 reviews from top 9 most frequent reviewers in Illinois. Table~\ref{tab:dataset_stats} summarizes both datasets in terms of authors, documents per author (D/A), words per sentence (W/S), and sentences per document (S/D).

\begin{table}[!t]
\caption{Dataset Statistics for \textsc{Author10} and \textsc{Illinois9}}
\label{tab:dataset_stats}
\centering
\footnotesize
\setlength{\tabcolsep}{3pt}
\begin{tabular}{lccccc}
\toprule
\textbf{Dataset} & \textbf{Auth.} & \textbf{Docs} & \textbf{D/A} & \textbf{W/S} & \textbf{S/D} \\
\midrule
Author10  & 10 & 15070 & 1507 & 14.3 ($\pm$ 12.8) & 4.74 ($\pm$ 4.56) \\
Illinois9 &  9 &  3959 &  440 &  9.19 ($\pm$ 4.82) & 2.24 ($\pm$ 1.89) \\
\bottomrule
\end{tabular}
\end{table}

We compare against three DP-based variants from \cite{meisenbacher2024thinkingoutsidedifferentialprivacy}: (i) \textit{DP}, strict DP-Prompt with privacy budgets $\epsilon \in \{25, 100, 250\}$; (ii) \textit{Quasi-DP}, omitting logit clipping but retaining temperature-based sampling; and (iii) \textit{Non-DP}, paraphrasing without DP constraints. Additionally, we include \textsc{ALISON}~\cite{xing2024alisonfasteffectivestylometric}, a stylometry-grounded non-DP baseline.

\subsection{Evaluation Metrics}

\noindent\textbf{Utility metrics.}
We evaluate utility using three complementary metrics: (1) cosine similarity (CS): As in \cite{meisenbacher2024thinkingoutsidedifferentialprivacy}, we embed each text with three pre-trained models \textsc{all-minilm-l6-v2}, \textsc{all-mpnet-base-v2}, and \textsc{gte-small} \cite{li2023towards} and report the averaged \emph{cosine similarity (CS)} across the models; (2) perplexity (PPL) \cite{10.1145/3485447.3512232}, measured with GPT-2 \cite{radford2019language} to assess fluency; and (3) weighted KL divergence, which quantifies information loss by assigning greater importance to rare, informative tokens:
\begin{equation}
D_{\mathrm{W\text{-}KL}}(P\parallel Q) = \sum_{t\in V}\mathrm{IDF}(t)\,P(t)\,\log\frac{P(t)}{Q(t)},
\label{eq:wkl}
\end{equation}
where $P$ and $Q$ are term-frequency distributions and $\mathrm{IDF}(t)$ weights rare tokens. While cosine similarity may remain high despite dropping rare words (e.g., named entities, domain-specific terms), weighted KL divergence penalizes such losses. This is crucial for ensuring anonymized texts retain informational content, not just surface-level semantic similarity. [See supplementary material for more details.]

\noindent\textbf{Privacy metrics.}
We assess privacy using four complementary metrics: (1) BLEU \cite{papineni2002bleu}, to measure lexical divergence between original and anonymized texts; (2) authorship attribution F1, which quantifies how effectively the anonymized text conceals the original author's identity. A lower F1 indicates stronger privacy. Following standard adversarial setups, authorship classifiers are trained on original data and evaluated on anonymized data, utilizing DeBERTa-v3 \cite{he2021debertav3} for \textsc{Author10} and BERT \cite{devlin2018bert} (proven effective on short texts) for \textsc{Illinois9}; (3) relative gain ($\gamma$) \cite{mattern2022limits,meisenbacher2024thinkingoutsidedifferentialprivacy}, which highlights privacy-utility trade-off:
\begin{equation}
\gamma = \frac{S_p}{S_o} - \frac{A_p}{A_o},
\label{eq:gain}
\end{equation}
where $A$ and $S$ denote authorship F1 and cosine similarity, respectively, on original (subscript $o$) and anonymized (subscript $p$) data. Higher $\gamma$ indicates a better privacy--utility trade-off. (4) fluency-aware gain, which extends $\gamma$ by penalizing perplexity degradation:
\begin{equation}
\gamma_f = \frac{1}{2}\left(\frac{S_p}{S_o} + \min\left(\frac{P_o}{P_p}, 1\right)\right) - \frac{A_p}{A_o},
\label{eq:gain_fluency}
\end{equation}
where $P$ denotes perplexity. This metric prevents methods from improving privacy by severely degrading readability.

\subsection{Parameter Settings}

We use \textsc{Llama-3.2-3B-Instruct} \cite{touvron2023llama} as the base LLM for both profiling and rewriting, chosen for its instruction-following ability and open-source availability. We also evaluate \textsc{MiniCPM3-4B} \cite{hu2024minicpm} to test model agnosticism. For style profiling, we use $K=5$ samples per author, which we found sufficient empirically (see supplementary material for analysis over $K \in \{2,5,10,20\}$).

\section{Results and Discussion}
\label{sec:results}

\begin{table*}[!t]
\caption{Comparison with baseline anonymization methods on \textsc{Author10} and \textsc{Illinois9}.}
\label{tab:Mainresults}
\centering
\begin{threeparttable}
\footnotesize
\setlength{\tabcolsep}{1.75pt}
\begin{tabular}{lcc|ccc|ccc|ccc|c|cc}
\toprule
\multicolumn{2}{c}{} & \textbf{Original Text} &
\multicolumn{3}{c|}{\textbf{DP ($\varepsilon$)}} &
\multicolumn{3}{c|}{\textbf{Quasi-DP ($\varepsilon$)}} &
\multicolumn{3}{c|}{\textbf{Non-DP ($k$)}} &
\textbf{ALISON} &
\multicolumn{2}{c}{\textbf{Ours}} \\
\cmidrule(lr){4-6}\cmidrule(lr){7-9}\cmidrule(lr){10-12}\cmidrule(lr){13-13}\cmidrule(lr){14-15}
 & $\varepsilon$ / $k$ &
 & 25 & 100 & 250
 & 25 & 100 & 250
 & 50 & 10 & 3
 & -- & \textsc{LLama} & \textsc{MiniCPM} \\
\midrule
\multirow{6}{*}{\textbf{Author10}}
& CS $\uparrow$              & 1     & 0.589  & 0.812  & \underline{0.832}  & 0.347  & 0.810  & \textbf{0.833}  & 0.710  & 0.750  & 0.787  & 0.704   & 0.702 & 0.820 \\
& BLEU $\downarrow$          & 1     & 0.077  & 0.123  & 0.153  & \textbf{0.001}  & 0.121  & 0.153  & 0.049  & 0.063  & 0.088  & 0.094   & \underline{0.023} & 0.213 \\
& PPL $\downarrow$           & 41    & 8770   & 928    & 905    & 16926  & 982    & 925    & 816    & 1080   & 837    & 368.30  & \textbf{42.47} & \underline{61.53} \\
& Author F1 $\downarrow$     & 66.45 & \underline{7.13}   & 58.10  & 60.60  & \textbf{6.59}   & 57.84  & 61.13  & 46.83  & 49.88  & 53.10  & 29.53   & 26.02 & 49.46 \\
& Relative Gain ($\gamma$)   & --    & \textbf{0.482}  & -0.062 & -0.080 & 0.248  & -0.060 & -0.087 & 0.005  & -0.001 & -0.012 & 0.260   & \underline{0.310} & 0.076 \\
& Fluency-aware Gain ($\gamma_{\mathrm{flu}}$) & -- & \underline{0.190} & -0.446 & -0.473 & 0.076 & -0.445 & -0.481 & -0.325 & -0.357 & -0.381 & -0.037 & \textbf{0.442} & -0.001 \\
\midrule
\multirow{6}{*}{\textbf{Illinois9}}
& CS $\uparrow$              & 1     & 0.592  & 0.894  & \underline{0.914}  & 0.595  & 0.892  & \textbf{0.916}  & 0.812  & 0.840  & 0.879  & 0.606   & 0.709 & 0.890 \\
& BLEU $\downarrow$          & 1     & \textbf{0.013}  & 0.432  & 0.497  & \underline{0.015}  & 0.424  & 0.520  & 0.255  & 0.292  & 0.373  & 0.094   & 0.022 & 0.310 \\
& PPL $\downarrow$           & 98.83 & 220.66 & 89.15  & 96.99  & 222.16 & 89.63  & 94.93  & 82.19  & 75.89  & 84.72  & 461.37  & \underline{53.36} & \textbf{43.05} \\
& Author F1 $\downarrow$     & 76.78 & 23.42  & 61.73  & 64.86  & \underline{21.84}  & 59.90  & 65.35  & 49.32  & 51.22  & 54.94  & 29.72   & \textbf{20.76} & 47.90 \\
& Relative Gain ($\gamma$)   & --    & 0.287  & 0.090  & 0.069  & \underline{0.311}  & 0.112  & 0.065  & 0.170  & 0.173  & 0.163  & 0.220   & \textbf{0.439} & 0.266 \\
& Fluency-aware Gain ($\gamma_{\mathrm{f}}$) & -- & 0.215 & 0.143 & 0.112 & 0.235 & 0.166 & 0.107 & 0.264 & 0.253 & 0.224 & 0.023 & \textbf{0.584} & \underline{0.321} \\
\bottomrule
\end{tabular}

\begin{tablenotes}[flushleft]
\footnotesize
\item \emph{Note:} Best results are in \textbf{bold}; second-best are \underline{underlined}. Higher $\gamma$ and $\gamma_{\mathrm{f}}$ indicate better privacy--utility trade-offs.
\end{tablenotes}
\end{threeparttable}
\end{table*}

\textbf{Privacy-Utility Trade-off.} Across both datasets, our style-guided rewriting approach reduces authorship attribution F1 by 60--70\% relative to the original text. On \textsc{Author10}, authorship F1 drops from 66.45 to 26.02, substantially outperforming the strongest non-DP baseline, \textsc{Alison} (F1=29.53), and non-DP paraphrasing (F1=53.10). While \textsc{Alison} preserves semantic similarity (CS $\approx$ 0.70), it suffers from significantly higher perplexity (368.30) compared to our method (42.47), indicating that our profile-based generation yields more fluent text.
 Strict DP ($\varepsilon=25$) achieves high privacy (F1 $\approx$ 7.13) but at a catastrophic utility cost, with perplexity exploding to $\approx$ 8,770. In contrast, our approach maintains near-original readability (PPL 42.47 vs. original 41). 
Table~\ref{tab:Mainresults} also demonstrates the framework's adaptability across different base LLMs. While \textsc{Llama} performs more aggressive stylistic neutralization (yielding lower Authorship-F1 scores), \textsc{MiniCPM} leans towards stronger surface-level content preservation (yielding higher CS). Crucially, both models substantially reduce the authorship signal compared to the original text and the baselines. This indicates that the style-guided rewriting approach is robust and its overall effectiveness is not strictly bound to a single underlying architecture.

\subsection{Full \textit{vs.} Single-Dimension Style Profiles}
\label{sec:profile_ablate}

All earlier results rely on the default configuration where the LLM produces a \emph{Full} profile based on four features: tone, vocabulary choice, length, and punctuation. To measure the impact of each cue, we repeat exactly the same procedure with four single-dimension profiles (\emph{Tone}, \emph{Length}, \emph{Vocab}, and \emph{Punc}). 
\begin{table}[t]
\centering
\small
\caption{Ablation on style channels (Illinois9).  
See the supplementary document for full results.}
\begin{tabular}{l|ccc}
\toprule
\textbf{Metric} & \textbf{Baseline} & \textbf{Full Profile} & \textbf{Length} \\
\midrule
CS $\uparrow$             & 1.000 & 0.709 & 0.713 \\
BLEU $\downarrow$         & 1.000 & 0.022 & 0.022 \\
PPL $\downarrow$          & 98.83 & 53.36 & 54.71 \\
Author F1 (s) $\downarrow$& 76.78 & 20.76 & 19.32 \\
Gain ($\gamma$)           & --    & 0.439 & 0.461 \\
\bottomrule
\end{tabular}
\label{tab:ablation-illinois9}
\end{table}
As Table~\ref{tab:ablation-illinois9} shows, the \emph{Full} profile delivers the best overall privacy–utility trade-off. However, the \emph{Length}-only variant comes surprisingly close for \textsc{Illinois9}. It matches or slightly exceeds the \emph{Full} profile in privacy gain while preserving marginally higher cosine similarity. We attribute this to the short, structurally uniform nature of reviews, where sentence length is already a strong authorial fingerprint. On the more varied \textsc{Author10} corpus the gap narrows: \emph{Tone}, \emph{Length}, and even \emph{Vocab} come within 0.2--0.3 pt of the Full profile on both F1 and CS. While no single cue dominates across datasets, the Full profile remains the most \emph{stable} choice for the privacy–utility trade-offs.

\subsection{Style-Guided, Semi-Guided, and Unguided Rewrites}
\label{sec:guided_main}

\begin{table}[!t]
\caption{Rewrite Modes on \textsc{Author10} and \textsc{Illinois9}.}
\label{tab:guided-vs-semi-guided}
\centering
\footnotesize
\setlength{\tabcolsep}{4pt}
\begin{tabular*}{\linewidth}{@{\extracolsep{\fill}} l c c c @{}}
\toprule
      & \textbf{\shortstack{Style-\\Guided}}
      & \textbf{\shortstack{Semi-\\Guided}}
      & \textbf{Paraphrase} \\
\midrule
\textbf{Author10}               &       &       &        \\
CS $\uparrow$                     & 0.702 & 0.694 & 0.752  \\
BLEU $\downarrow$                 & 0.023 & 0.018 & 0.034  \\
PPL $\downarrow$                  & 42.47 & 42.39 & 56.04  \\
Author F1 $\downarrow$            & 26.02 & 24.10 & 36.61  \\
Weighted KL $\downarrow$ (All)    & 48.75 & 49.48 & 41.11  \\
Weighted KL $\downarrow$ (Subset) & 39.39 & 42.96 & 33.60  \\
\addlinespace
\textbf{Illinois9}              &       &       &        \\
CS $\uparrow$                     & 0.709 & 0.746 & 0.791  \\
BLEU $\downarrow$                 & 0.022 & 0.027 & 0.039  \\
PPL $\downarrow$                  & 53.36 & 52.85 & 61.46  \\
Author F1 $\downarrow$            & 20.76 & 22.64 & 29.37  \\
Weighted KL $\downarrow$ (All)    & 58.33 & 57.16 & 46.87  \\
Weighted KL $\downarrow$ (Subset) & 52.10 & 49.97 & 41.59  \\
\bottomrule
\end{tabular*}
\end{table}

To evaluate the impact of explicit style profiles on anonymization performance, Table~\ref{tab:guided-vs-semi-guided} shows our default \textit{style-guided} rewriting mode against the \textit{semi-guided} and \textit{unguided} (paraphrase) baselines. On \textsc{Author10} the \textit{semi-guided} setting achieves slightly better Author-F1 scores but at the cost of noticeably worse utility: \emph{PPL} rises, \emph{CS} falls, and \emph{weighted KL divergence} increases, indicating greater distortion of informative tokens. Further analyses in the Supplementary Material (LLM-as-judge and named-entity retention experiments) confirm that our \textit{style-guided} rewrite maintains more informational content while achieving nearly the same privacy levels as the \textit{semi-guided} variant. Taken together, these patterns also hint that the authorship classifier is not relying solely on stylistic cues but also draws on residual content when identifying the author. A fully \textit{unguided}/\textit{paraphrase} version does even worse, yielding the largest utility loss while leaving a sizable authorship signal, underscoring the value of explicit style guidance.

\subsection{Qualitative Insights and Robustness}

\begin{table}[!t]
\caption{Illustrative long-form rewrite example from \textsc{Author10} (excerpted for space).}
\label{tab:qualitative_case}
\centering
\footnotesize
\setlength{\tabcolsep}{2pt}
\renewcommand{\arraystretch}{0.95}
\begin{tabularx}{\linewidth}{@{}lXcc@{}}
\toprule
\textbf{Method} & \textbf{Text (excerpt)} & \textbf{Identified?} & \textbf{W-KL} \\
\midrule
Original & I read an interesting article on human stupidity... This post continues my earlier discussion. I am also looking out for new designs for this blog and urlLink QSS. Anyone having any good color schemes, send them to me. & Yes & -- \\
DP\\(\(\varepsilon=100\)) & I am looking for new designs for this blog and urlLink QSS. & No & 48.62 \\
Non-DP\\(\(k=25\)) & I am looking for new designs for this blog and urlLink QSS. & Yes & 48.62 \\
Ours\\(style-guided) & I recently came across an intriguing piece about human behavior patterns... This post serves as a follow-up to my previous discussion on the topic... & No & 43.54 \\
\bottomrule
\end{tabularx}
\end{table}

The central dilemma of text anonymization is removing stylistic fingerprints without erasing essential content. Table~\ref{tab:qualitative_case} illustrates this trade-off with a representative long-form example from \textsc{Author10}. The original post contains multiple discourse elements: discussing an article, continuing an earlier thread, and asking for blog design suggestions. 
The DP rewrite ($\varepsilon=100$) avoids attribution but drastically collapses the text into a single sentence, discarding the majority of the original meaning. The Non-DP ($k=25$) baseline produces a similarly truncated sentence but still fails to mask the author, demonstrating that severe lexical reduction alone does not guarantee privacy. In contrast, our style-guided rewrite successfully avoids identification while preserving the full discourse structure and intent. This superior retention of informative lexical content is further reflected by its lower W-KL divergence (43.54 vs. 48.62 for the baselines).

Extensive robustness checks are detailed in the supplementary material. These include evaluations against a method-aware, white-box adversary who retrains the authorship classifier on anonymized data; our approach still substantially weakens stylometric cues, reducing attacker F1 by 55\% (from 77 to 35) on \textsc{Illinois9}. We also verify out-of-domain generalization on Yelp and IMDB datasets, where authorship F1 similarly drops by more than 85\%.

\section{Conclusion}
We present a text anonymization framework that employs explicit, prompt-driven stylistic profiling to effectively mask author identity while preserving semantic content. Our approach demonstrates robust performance across both long-form and short-form text datasets, significantly outperforming differential privacy-based and paraphrasing baselines. Evaluations show that our style-guided anonymization better retains original content and meaning compared to noise-based approaches, and remains effective even against a method-aware white-box adversary. These results highlight the critical role of precise stylistic manipulation in achieving a favorable privacy--utility trade-off. 

\noindent\textbf{Limitations and Future Work.} While promising, our framework presents several avenues for future work. First, it relies on the underlying LLM's capacity to extract predefined style features, potentially missing nuanced rhetorical patterns or syntactic idiosyncrasies that serve as authorial fingerprints. Future research should explore unsupervised style-profiling to reduce reliance on hand-picked cues. Second, standard authorship F1 metrics can conflate style obfuscation with content loss, necessitating new evaluation frameworks that disentangle these factors. Third, our experiments focus on text corpora; extending evaluation to ASR-transcribed text would clarify how stylometric leakage interacts with acoustic anonymization pipelines. Finally, extending this approach to address user fairness and multimodal contexts (e.g., timestamps or geolocation metadata) remains a critical next step for real-world deployment.



\section{Generative AI Use Disclosure}
During the preparation of this manuscript, the authors utilized generative AI tools to assist in editing, condensing, and polishing the text. The authors comprehensively reviewed and edited all AI-generated suggestions, take full responsibility for the content of the publication, and confirm that generative AI was not used to produce the core scientific contributions, methodology, or results.

\section{Acknowledgements}

The authors gratefully acknowledge the RMIT Advanced Computing Ecosystem (RACE) for providing computational resources and technical support for this research.

\bibliographystyle{IEEEtran}
\bibliography{anthology, mybib}

\end{document}